\documentclass{article}

\usepackage{arxiv}

\usepackage[utf8]{inputenc} 
\usepackage[T1]{fontenc}    
\usepackage{hyperref}       
\usepackage{url}            
\usepackage{booktabs}       
\usepackage{amsfonts}       
\usepackage{nicefrac}       
\usepackage{microtype}      
\usepackage{cleveref}       
\usepackage{lipsum}         
\usepackage{graphicx}
\usepackage{natbib}
\usepackage{doi}

\title{eXplainable Artificial Intelligence (XAI) in aging clock models}

\author{ \href{https://orcid.org/0000-0001-9277-502X}{\includegraphics[scale=0.06]{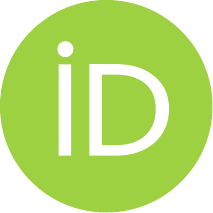}\hspace{1mm}Alena Kalyakulina}\\
	\texttt{kalyakulina.alena@gmail.com} \\
	\And
	\href{https://orcid.org/0000-0002-0540-9281}{\includegraphics[scale=0.06]{orcid.pdf}\hspace{1mm}Igor Yusipov} \\
	\texttt{yusipov.igor@gmail.com} \\
	\And
	\href{https://orcid.org/0000-0002-3248-1633}{\includegraphics[scale=0.06]{orcid.pdf}\hspace{1mm}Alexey Moskalev} \\
	\texttt{amoskalev@list.ru} \\
	\And
	\href{https://orcid.org/0000-0001-9841-6386}{\includegraphics[scale=0.06]{orcid.pdf}\hspace{1mm}Claudio Franceschi} \\
	\texttt{claudio.franceschi@unibo.it} \\
	\And
	\href{https://orcid.org/0000-0002-1903-7423}{\includegraphics[scale=0.06]{orcid.pdf}\hspace{1mm}Mikhail Ivanchenko} \\
	\texttt{ivanchenko.mv@gmail.com} \\
}

\renewcommand{\shorttitle}{XAI in age prediction}

\hypersetup{
pdftitle={XAI in age prediction},
pdfsubject={cs.AI},
pdfauthor={Alena Kalyakulina, Igor Yusipov, Maria Giulia Bacalini, Alexey Moskalev, Claudio Franceschi, Mikhail Ivanchenko},
pdfkeywords={explainable artificial intelligence, aging, longevity, age-related diseases, machine learning},
}

\begin{document}
\maketitle

\begin{abstract}
	eXplainable Artificial Intelligence (XAI) is a rapidly progressing field of machine learning, aiming to unravel the predictions of complex models. XAI is especially required in sensitive applications, e.g. in health care, when diagnosis, recommendations and treatment choices might rely on the decisions made by artificial intelligence systems. AI approaches have become widely used in aging research as well, in particular, in developing biological clock models and identifying biomarkers of aging and age-related diseases. However, the potential of XAI here awaits to be fully appreciated.
	
	We discuss the application of XAI for developing the "aging clocks" and present a comprehensive analysis of the literature categorized by the focus on particular physiological systems.
\end{abstract}


\section{Introduction}\label{sec:introduction}
Machine learning (ML), and deep learning (DL) in particular, is currently one of the most common data analysis approaches in applications. Deep models handle large amounts of input data, training many layers, but in most cases, their functioning is not transparent. In this regard they are often called black boxes \citep{Saleem2022}. Decision-making process in such deep architectures is difficult to explain, raising concerns about the trustworthiness of such models and the security of their deployment. The problem of explainability of artificial intelligence (AI) models has received much attention \citep{Baehrens2010, Lipton2018, Samek2017, Simonyan2014}, and made eXplainable Artificial Intelligence (XAI) an important area of AI \citep{Nauta2023}. Major goals of XAI are to develop approaches capable of uncovering the grounds behind model decision-making, and, more profoundly, to develop interpretable and logically explainable models. XAI explanations must be understandable, reliable, whereas the explained models must retain predictive accuracy \citep{Saleem2022}. Commonly, one distinguishes global and local explainability. Global interpretation allows one to ‘open’ the black box of AI models by explaining the predictions of the model as a whole. Local interpretation explains the model's decision for each particular sample. These two types of explainability represent two sides of the same coin: global explainability allows one to establish general patterns, while local explainability tracks these patterns at the individual level. The lack of explainability significantly limits the use and deployment of models, especially in sensitive applications, where human life and health may depend on ML decision.

In medical applications AI based clinical decision support systems aim to assist clinicians in diagnosing diseases and making treatment decisions \citep{Amann2020}. One of the most important requirements for such systems is clinical validation and the ability to verify model decisions. Biomedical data is subject to various errors (recording and processing errors, natural noise, and others), so AI systems will inevitably make errors as well. Another source of errors may be systemic problems caused, in particular, by limited train subset. In this case, AI models may make errors because an individual sample may significantly differ from the set on which the model was trained \citep{Amann2020}. In any case, explainability is extremely important; it allows one to track different types of errors, identify their causes, and adjust the behavior of the system. Explainability can also offer a personalized approach based on individual patient characteristics and risk factors. XAI approaches can give an explanation in natural language or visualize how different factors affected the final outcome (risk score, diagnosis, or proposed treatment), which can increase patient awareness with proper use \citep{Politi2013, Stacey2017}, and allow clinicians to make confident clinical decisions, adapting predictions and recommendations to individual circumstances if necessary \citep{Beil2019}.

A central problem of geroscience is identifying biomarkers of aging, which has benefited much from the growth of ML and DL research \citep{Moskalev2023, Zhavoronkov2019a}. Considerable attention here is devoted to estimating individual biological age, which would characterize personal health status in various aspects, being, in general, different from chronological age. This biologically meaningful characteristic is associated with risk of mortality, disease, and general well-being, and even though each individual feature employed in one or another mathematical estimator of biological age may not be explicitly related to age, a combination of features are known to have predictive power \citep{Zhavoronkov2019}. Various types of data may be used to estimate age, including laboratory tests, magnetic resonance (MRI) and X-ray images, electrocardiogram (ECG) and electroencephalogram (EEG) signals. Development and analysis of age estimators can help, in particular, in the study of age-related diseases \citep{Zhavoronkov2011}, immunological aging, response to medications and vaccines \citep{Zhavoronkov2019} and in many other healthcare applications. The recent promises on the possibility of extending life expectancy \citep{Partridge2018} demand reliable models that estimate biological age and the aging rate to validate life extension techniques \citep{Rutledge2022}. Age estimation models are usually called “biological” or “aging clocks”. One of the first fundamental works in this field are the epigenetic clocks from Hannum \citep{Hannum2013} and Horvath \citep{Horvath2013}, which estimates age using a linear model based on DNA methylation data. By now a plethora of models have been proposed employing different input data and implementing AI techniques \citep{Mamoshina2019, Zhavoronkov2019}. There are several challenges in overcoming which XAI can be helpful. In particular, it is not always possible to separate the chronological and biological aging components in age estimation models \citep{Bell2019}. Here, the global aspect of XAI approaches can be used to identify hallmarks for particular age stages in healthy individuals that are part of the chronological aging component, as well as the effects of the environment, medical history, and many other factors that contribute to variations in the biological age component. The XAI can also help with identifying the set of biomarkers that are most representative of an individual's health status, as well as discarding redundant ones.

\begin{figure}
	\centering
	\includegraphics[width=0.99\textwidth]{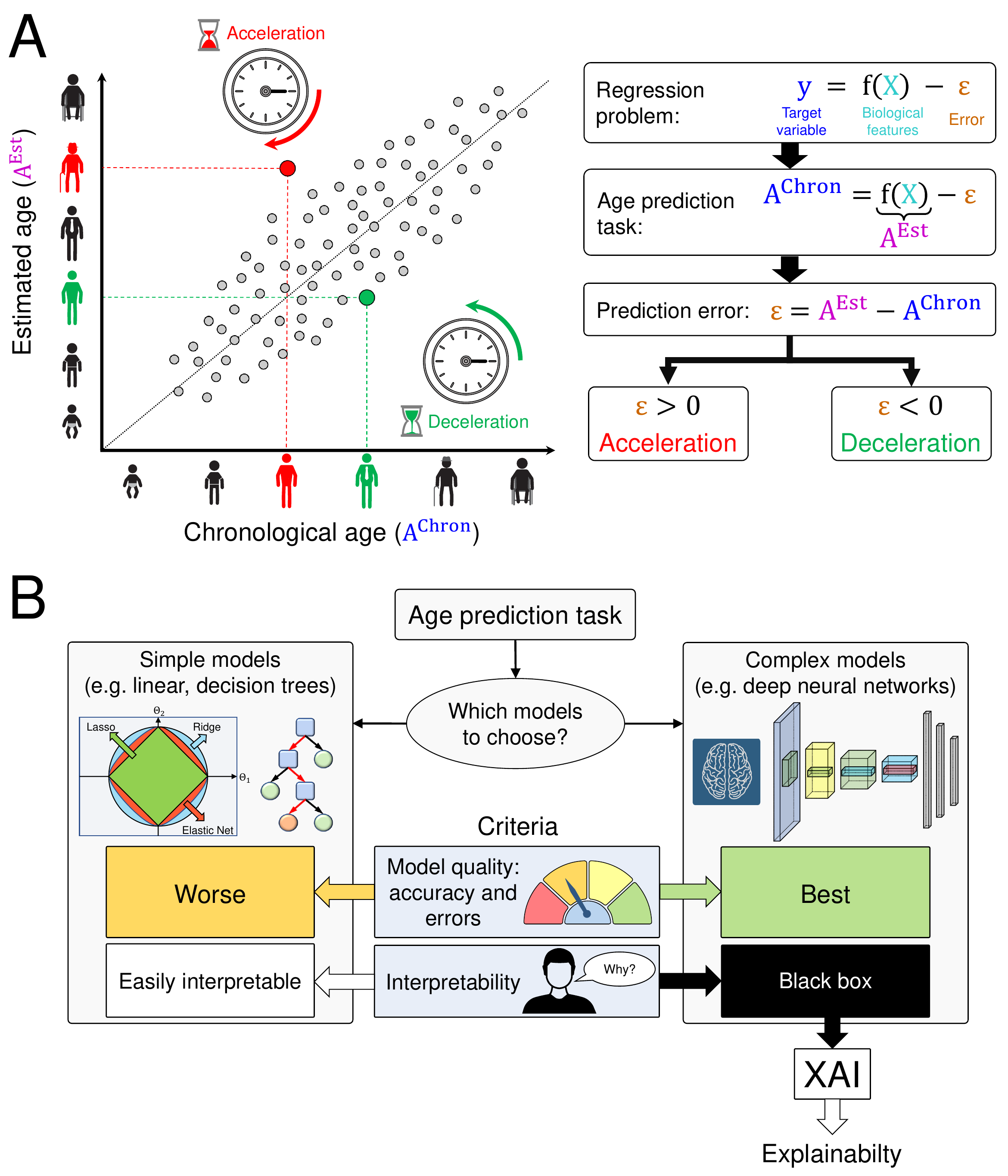}
	\caption{(A) Chronological and estimated age in the age prediction task. ML models seek to predict chronological age ($A^{Chron}$) with as small error ($\varepsilon$) as possible based on information on various biomarkers, resulting in estimated age ($A^{Est}$), which is usually used as an estimate of the biological age. If a person's estimated age is higher than chronological age, it is named ‘age acceleration’ (this is usually interpreted as a person's health being worse than average for their age). If a person's estimated age is lower than chronological age, it is named ‘age deceleration’ (this is usually interpreted as a person's health being better than average for their age). (B) Model selection for solving the age prediction task. Models can be conventionally divided into simple (classical ML approaches) and complex (DNNs). The first important criterion in model selection is its quality, namely the ability to predict age as accurately as possible. Usually, simple models have worse results than DNN architectures. The second important criterion is interpretability, namely the ability to obtain explanations of why the model makes these decisions. Simple models are usually easy to interpret, while complex models are ‘black boxes’ and require XAI methods to ‘open’ them.}
	\label{fig:age_prediction}
\end{figure}

The considered points are summarized in Figure \ref{fig:age_prediction}. Age estimation is commonly a regression problem that can take various biological features as an input. Models seek to predict a person's chronological age with as small an error as possible. Commonly, one uses the model's output as an estimate of the biological age (ideal estimation of person's health status), and the mismatch between chronological age and biological age is usually treated as age acceleration (if biologically the person is older than chronologically) or age deceleration (if biologically the person is younger than chronologically), as shown in Figure \ref{fig:age_prediction}A. It has to be kept in mind that the difference can be due to many reasons, like imperfection of the model, heterogeneity between individuals (not related to biological aging), actual difference between biological and chronological age and others. The problem of age estimation inevitably raises the question of choosing a suitable model. Among the many criteria that influence the final choice, there are two main ones: model performance in terms of predictive accuracy and interpretability (Figure \ref{fig:age_prediction}B). Classical, simple models (e.g., linear, treelike ones) are usually easy to interpret, but they may not take into account the complex relationships between the input features, thereby showing worse results. More complex state-of-the-art models (like deep neural network (DNN) architectures) often give better results, but they are black boxes, which do not allow for explaining the principles by which models make decisions. This is where XAI approaches come in.

The primary aim of this review is to provide the current state-of-the art of XAI methods applied to human biological age estimation. It has the following structure. Section \ref{sec:AI} describes the age estimation problem and XAI methods in the context of ML approaches, and gives an overview of XAI methods and their application to different types of data. Section \ref{sec:XAI} presents a comprehensive review of studies that use XAI methods to explain the results of age estimation models, grouped by the assessed body systems and the types of input data used. Section \ref{sec:conclusion} concludes the work, summarizing the achievements of XAI in the field of age-associated changes in the human body.

\section{Artificial Intelligence, data types and methods}\label{sec:AI}

\begin{figure}
	\centering
	\includegraphics[width=0.99\textwidth]{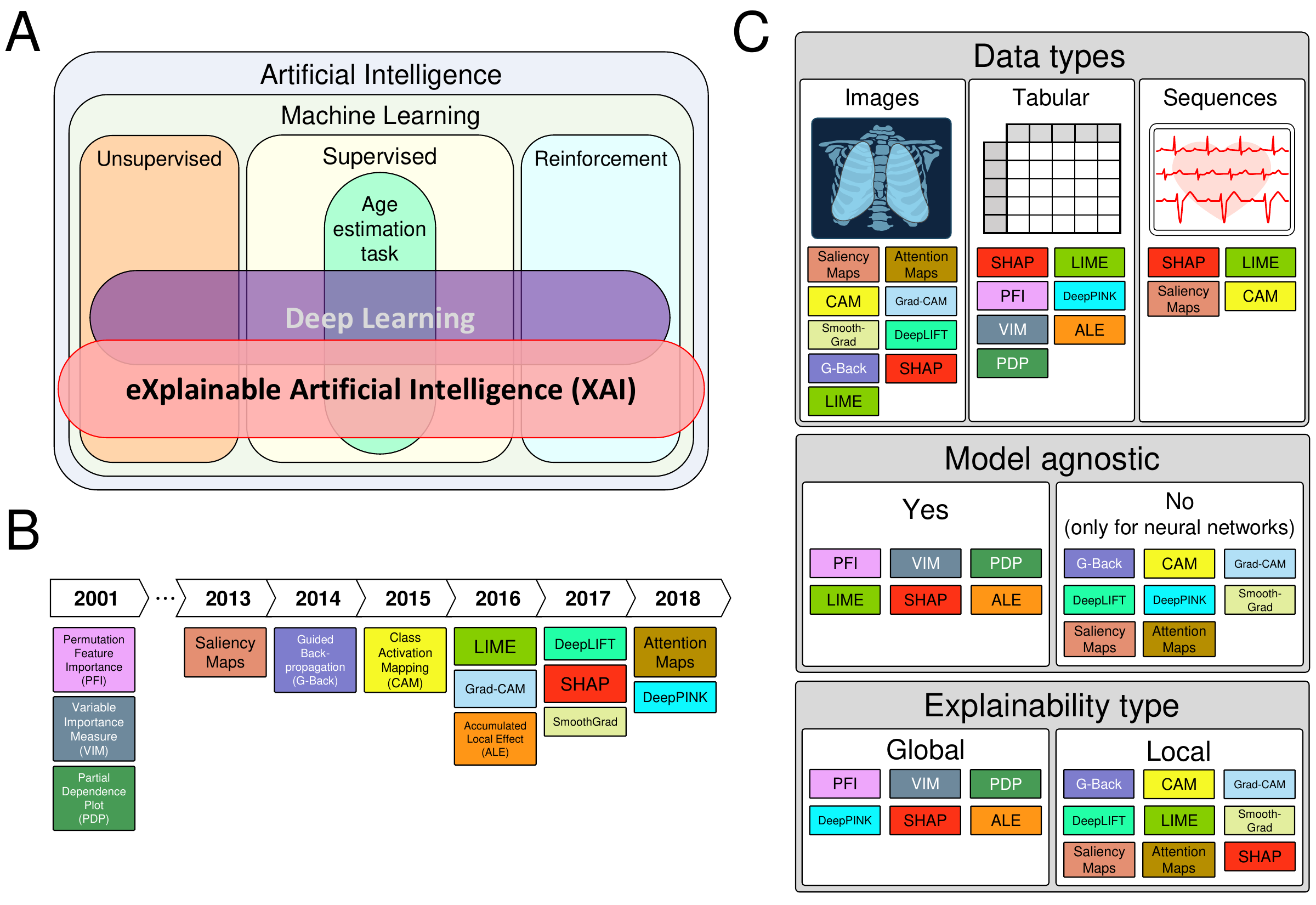}
	\caption{XAI methods used in aging clock models. (A) Schematic representation of the age estimation problem location on the general scheme of AI areas, including XAI methods. (B) Timeline of development of XAI methods mentioned in this review. Dates are determined by the first presentation of methods in the public domain (the first version of the preprint or article). (C) Different types of classifications of the considered XAI approaches: depending on the type of input data (image, table, sequence), model-agnostic or model-specific, explainability type (global or local).}
	\label{fig:AI}
\end{figure}

AI is an extremely broad field of research and development, finding its application in almost all spheres of modern life and helping to solve many complex problems. AI makes it possible to create data-driven decision-making systems \citep{Ali2023}. ML models are a broad class of AI approaches, combining both classical models and more complex DNNs. The main types of ML are supervised learning, unsupervised learning, and reinforcement learning. Depending on the type of the predicted value in supervised learning, there are classification tasks (class is predicted) and regression tasks (continuous value is predicted). Biological age estimation is usually done as a chronological age regression task (age - continuous value - is predicted) on different biomedical data, cf. Figure \ref{fig:AI}A. The formulation as an age range classification task is much less common. Biological age can be estimated using both simple methods and complex neural network architectures. At the same time, an increasing amount of data requires more and more advanced methods to be able to detect nonlinear relationships between input parameters and their impact on the final result of the biological age estimation. Most neural network architectures are ‘black boxes’, the models with unclear routes to decisions and unpredictable consequences to which changes in input data may lead. This is where XAI methods can help.

The timeline of the development of XAI methods mentioned in this review is shown in Figure \ref{fig:AI}B. It started with the permutation feature importance (PFI), variable importance measure (VIM) \citep{Breiman2001}, and partial dependence plot (PDP) \citep{Friedman2001} proposed in 2001. PFI, one of the types of VIMs, is used for tabular data and represents a reduction in the model estimate when a single feature value is randomly shuffled. These metrics were originally proposed for tree models, but can also be used for other methods. PDPs are used to analyze and visualize the interaction between the set of interesting input features and the final model estimate. One of the first approaches developed for input image data is saliency maps \citep{Simonyan2014}, which are still widely used and have been adapted for sequence input data as well. Saliency maps show which regions of the input images were used by the AI model to make decisions and can provide a visual representation of how regions important to the model fit with human attitudes. Guided Backpropagation \citep{Springenberg2015} builds on the ideas of Deconvolution \citep{Zeiler2013} and Saliency \citep{Simonyan2014}, solving the problem of negative gradient flux and minimizing the noise they cause. This approach is also used for input image data. Another well-known method for images and sequences is Class Activation Mapping (CAM) \citep{Zhou2016}. CAM usually uses a global average pooling layer after the convolutional layers and before the final fully connected layer. Later, a Grad-CAM (Gradient-weighted Class Activation Mapping) modification was proposed, which uses a gradient approach to generate CAM \citep{Selvaraju2020}. Accumulated local effects (ALE) \citep{Apley2020} are similar to the concept of PDPs in that they both aim to describe how functions on average affect model predictions. ALE eliminates the bias that occurs in PDPs when the feature of interest is highly correlated with other features. LIME (Local Interpretable Model-agnostic Explanations) provides locally accurate explanations in the neighborhood of the explained instance \citep{Ribeiro2016}. After obtaining a surrogate dataset, it weighs each row according to how close they are to the original sample and uses a feature selection method, such as Lasso, to obtain the most important features. As the name implies, this approach can be applied to any model and provides only local explainability. DeepLIFT (Deep Learning Important FeaTures) is a method of decomposing the output prediction of a neural network to a particular input by back propagating the contributions of all neurons in the network to each input feature, comparing the activation of each neuron to its ‘reference activation’ \citep{Shrikumar2017}. It is applied only to neural networks. One of the best known and most widely used XAI methods is SHAP (Shapley Additive exPlanations), a game-theoretic approach to explaining the results of any ML model. It relates optimal credit allocation to local explanations using classical Shapley values from game theory and related extensions \citep{Lundberg2017}. SHAP is applicable to almost any model, input data type, and is used for both global and local explainability. SmoothGrad \citep{Smilkov2017} adds Gaussian noise to the input data and calculates the average of all samples to reduce the importance of less frequent features. This approach is also specific to neural networks and is typically used for image input data. One of the most recent approaches for explaining the results of convolutional neural networks (CNNs) with input image data is attention maps \citep{Jetley2018}. This approach generates intermediate representations of the input image at different stages of the CNN pipeline and outputs a two-dimensional matrix of scores for each map. As for saliency maps, CAM and Grad-CAM, this approach allows for comparing the representation of the neural network and the human in terms of important regions of the image highlighted for decision making. DeepPINK (Deep feature selection using Paired-Input Nonlinear Knockoffs) \citep{Lu2018} offers a special pairwise-connected layer for the neural network to encourage competition between each original feature and its knockoff counterpart. DeepPINK offers an algorithm-independent measure of feature importance with more power than the naïve combination of the knockoffs idea with a vanilla multilayer perceptron.

ML models can be built on different types of data that they take as an input. Among the most common data types are images, tables, and sequences (like signals or texts). Images and sequences are examples of unstructured data, while for tabular data the features are already extracted. For different types of data, various types of methods dominate: for images and sequences, CNNs are the most common \citep{Hershey2017, Sultana2018}, for tabular data gradient-boosted decision trees (GBDTs) has long shown the best results, but now specialized neural network approaches that adapt techniques from other fields are also actively developing \citep{Borisov2022, Grinsztajn2022, Shwartz-Ziv2022}. XAI methods are also mostly specific to different types of data - some methods are applied to images and sequences, others are applied to tabular data (Figure \ref{fig:AI}C). But approaches like SHAP and LIME can be adapted to almost any type of input data and different models, and are very common in a wide range of applications. Different maps are usually used for images - saliency maps, CAM, attention maps and their modifications, as well as methods based on backpropagation. Classical approaches such as PFI, PDP and their modifications, as well as DeepPINK, are used for tabular data.

XAI methods are divided into model-agnostic, which can be applied to any type of ML models, and model-specific, which can be applied only to a certain class of models (usually to neural network architectures). Among model-agnostic methods there are such classical ones as PFI, VIM, PDP, ALE, as well as more modern SHAP and LIME. The other considered methods are specific for neural networks (usually convolutional ones). Another type of classification of XAI approaches includes global and local explainability. Global explainability attempts to interpret the behavior of the model as a whole, revealing general patterns. Local explainability attempts to obtain an interpretation for each individual sample and to identify the features that affect the decision in each case. The attribution of methods to different groups is shown in Figure \ref{fig:AI}C.

\section{XAI in age prediction studies}\label{sec:XAI}

\begin{figure}
	\centering
	\includegraphics[width=0.99\textwidth]{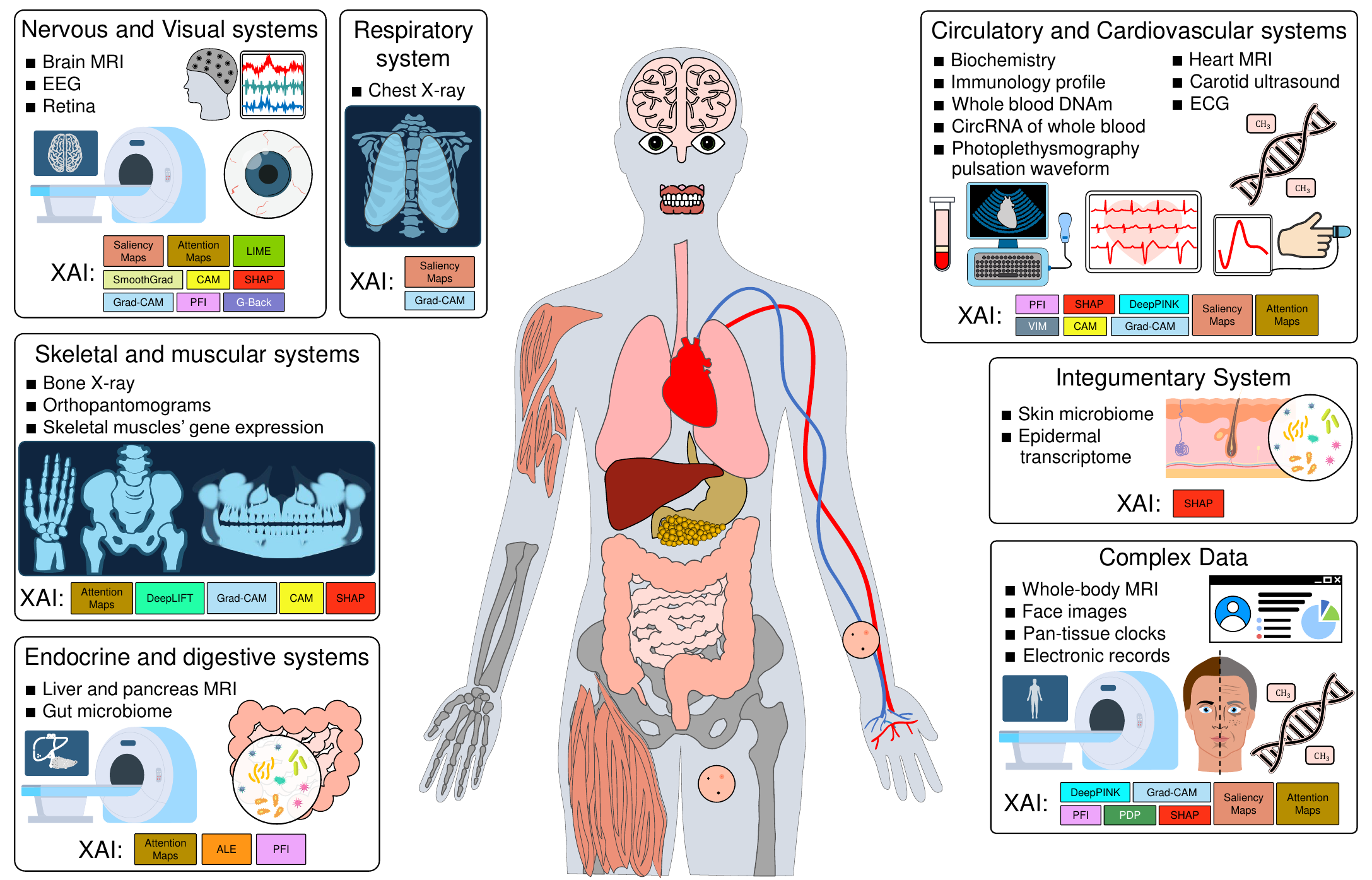}
	\caption{Body systems and corresponding biomarkers used for biological clocks that implement XAI. In the center: the human body with highlighted organs and tissues giving measurables as inputs for various age estimation models. Each individual box corresponds to one or two body systems and lists the main data types that are employed. The complex data block represents different complex sets of biomarkers related to multiple organism systems.
	}
	\label{fig:XAI}
\end{figure}

Among a wealth of age estimation models, some already implement XAI approaches to identify the most important features that contribute to the result. The models may address specific body systems or characterize an organism as a whole (Figure \ref{fig:XAI}). Correspondingly, they take in system- or tissue-specific data, or aggregate information from multiple body systems.

\subsection{Nervous and Visual Systems}\label{sec:XAI:subsec:nervous}

\subsubsection{Brain age estimation with MRI data}\label{sec:XAI:subsec:nervous:MRI}

Brain age estimation is one of the most common tasks due to the close relationship between brain function, aging process, and neurodegenerative age-related diseases \citep{Cole2017}. MRI is a common noninvasive test that is very informative in assessing the functional state of the brain. Higher brain age compared to chronological age is evidence of accelerated aging and requires attention with regard to the potential progression of neurological, neurodegenerative, and psychiatric diseases \citep{BoscoloGalazzo2022}, while lower brain age compared to chronological age is interpreted as decelerated aging and corresponds to a good health status. To estimate brain age, various metrics and morphological characteristics of MRI scans or whole images can be used as input for ML models. The most advanced DL approaches demand that XAI techniques are employed in order to understand which MRI parameters affect the final model solutions and in what way, to increase the confidence in the models, required in clinical applications.

Brain aging clocks were proposed even before XAI methods became widespread. However, the need to determine the importance of certain features and to explain model predictions was articulated even at that time. In particular, \citep{Cherubini2016} made one of the first attempts to visualize the most important MRI regions based on the analysis of individual voxels (single elements of a 3D brain image). Age estimation was performed by means of a simple linear model, where the weight coefficients for individual voxels allowed for a straightforward identification of the most important brain regions for age estimation on the 3D model.

The development of DL models led to architectures taking whole images as an input, like CNNs. Such models have also become widespread in the age estimation based on MRI images. However, the explanation here is significantly complicated by the large dimensionality and complexity of the model. Comparatively simple methods for explaining the importance of individual brain regions in age estimation are computationally costly, although they have been used quite successfully (e.g., PFI \citep{Kolbeinsson2020}). More advanced XAI methods have also been applied, relying on various maps that highlight the most active regions in the image: attention maps, CAM, saliency maps \citep{Feng2020, Hepp2021, Hu2021, Lam2020, Mouches2022, Ren2022, Wang2019, Yin2023}, SmoothGrad \citep{Levakov2020, Wilms2021}. These approaches have also uncovered interesting relationships between aging, brain function, and neurodegenerative diseases. In particular, it allowed to identify brain regions associated with the development of dementia \citep{Wang2019}, as well as those associated with both age and Alzheimer's disease \citep{Lam2020}. In \citep{Levakov2020}, the aggregation of explanation maps for many samples identified the association of cerebrospinal fluid (CSF) volume with age. Changes in the activation patterns of different brain regions with age were demonstrated \citep{Feng2020}. For younger participants, a smaller error was observed in (Hepp et al., 2021), and higher estimation accuracy in central brain regions was also shown. In \citep{Hu2021} it was found that the most age-associated brain regions in children and adolescents are associated with movements, language, and processing of sensory information like vision and sense of touch (precentral gyrus, postcentral gyrus, inferior parietal lobule, middle temporal gyrus, medio ventral occipital cortex). In contrast, in the elderly, the amygdala, hippocampus, and thalamus, which are responsible for the limbic system, behavioral and emotional responses, and consolidating memories, were the most involved in age estimation \citep{Ren2022}. Sex-specific aspects of brain aging were revealed in \citep{Yin2023}.

U-Net architecture-based importance map extraction approach and related modifications (U-Noise) is less popular, but nevertheless noteworthy \citep{Bintsi2021, Popescu2021}, as well as the guided-backpropagation approach \citep{Cho2022}.

Another representation of MRI data is tabular, which characterizes morphological features: the sizes and volumes of the main structural brain regions measured from MRI images, as well as the ratios between them. As the input data in this case are in the tabular form (with numeric, ordinal, or categorical values), and all features have already been extracted, XAI approaches to explaining age estimation models based on such data are different, typically, SHAP and LIME \citep{Ball2021, Ballester2023, Cumplido-Mayoral2023, Gomez-Ramirez2022, Han2022, Lombardi2021, Ran2022, Salih2021, Scheda2022}. They also allow to make conclusions about the influence of morphological features on brain age estimation. Interestingly, a direct comparison of these methods for local explainability of brain age estimation showed little agreement between the methods: SHAP method emphasized the importance of metrics related to the precentral gyrus, inferior and lateral occipital cortex, and statistical descriptors of CSF volume, whereas LIME method valued more white matter volumes of opercular and triangular part of inferior frontal gyrus and inferior temporal gyrus \citep{Lombardi2021}. In \citep{Ball2021} XAI was used for identifying typical changes during adulthood that appeared to reflect developmental remodeling in the cortex based on age estimation models for a cohort of children and adolescents. Similar observation was made in \citep{Scheda2022}: cortical thickness and brain complexity manifested maximum contribution to age estimation in both children and adults; like total intracranial volume and cortical thickness in \citep{Han2022} for different models. Brain-to-intracranial-volume ratio appeared to be the most important in estimating brain age \citep{Gomez-Ramirez2022}. The brain age vector, stroked with Shapley values, has been shown to be a useful tool for the early screening of mild cognitive impairment and Parkinson's disease \citep{Ran2022}. In \citep{Ballester2023} the relationship between brain age and total gray matter volume in schizophrenia was revealed using SHAP. The use of different brain MRI metrics in \citep{Cumplido-Mayoral2023} revealed the following most age-associated brain regions in both males and females: amygdala, nucleus accumbens, cerebellar white matter, lateral ventricles, and insula. Sex-specific traits were also found, for females: the thicknesses of the transverse temporal cortex, the pars triangularis, the inferior parietal cortex, and the left frontal pole, as well as the volume of the left entorhinal cortex; for males: the thicknesses of the left isthmus cingulate, the right cuneus, and the cortical volume of the superior frontal and right rostral middle regions. An unusual application of SHAP was proposed in \citep{Sun2022}: instead of morphological characteristics, voxel MRI parameters were used as input data, and SHAP highlighted important voxels, which were then used to build dynamic connectivity networks. 

An original approach was proposed in \citep{Monti2020}: functional connectivity networks were constructed for different brain regions, which were then used to solve the age prediction problem. Self-explainable normalizing flows were proposed in \citep{Wilms2021}, demonstrating comparable results with CNN model and SmoothGrad XAI approach.

\subsubsection{Brain age estimation with EEG data}\label{sec:XAI:subsec:nervous:EEG}

Another common non-invasive test to evaluate brain function and activity is EEG. Changes in the brain caused by aging or age-associated diseases can affect the electrophysiological activity of the brain. Either raw EEG signals or their frequency, spectral, amplitude, and/or other characteristics serve the input for ML models.

XAI approaches in the age estimation based on EEG data are not commonly used; more often, built-in mechanisms for determining the feature importance are employed \citep{AlZoubi2018, Sun2019, Vandenbosch2019}. Nevertheless, they are able to identify important EEG age-associated features. In particular, it was shown in \citep{Vandenbosch2019} that low-frequency power decreases sharply from childhood to adolescence, with peak alpha frequency increasing with age and peak alpha power decreasing with age. A study of EEG during sleep \citep{Sun2019} has shown that the duration of Non-REM (non-rapid eye movement) phases of sleep, as well as the total sleep time and the awakening time rate influence accelerated brain aging. In recent work \citep{Khayretdinova2022}, deep CNNs were used to estimate brain age, and attribution maps were used as an XAI method to identify the EEG signal elements whose activity most affects age estimation. Interestingly, the state of the eyes (open or closed), as well as the activity from the frontal electrodes (which may reflect eye movement activity) were found to be significant features.

\subsubsection{Retinal age estimation}\label{sec:XAI:subsec:nervous:retina}

Health changes, symptoms of various diseases, age-associated conditions can be reflected in the eye and, in particular, the retina, so the retinal age estimation seems to be a promising and interesting direction. Age is a basic risk factor in the development of many eye diseases, leading to a significant decrease in visual acuity and even to vision loss. It is also worth noting that most tests that assess retinal age are non-invasive and can be conveniently used in clinical applications, for example, to track early signs of diseases (particularly cardiovascular diseases) or to estimate the rate of their development. The types of data used for the age estimation include retinal fundus photographs \citep{Nusinovici2022, Poplin2018, Zhu2023}, anterior segment morphological features \citep{Ma2021}, and macular optical coherence tomography (OCT) \citep{Chueh2022, Shigueoka2021}.

For image data in retinal age estimation, one employs CNNs and corresponding XAI methods: attention maps \citep{Poplin2018, Zhu2023}, Grad-CAM \citep{Chueh2022, Shigueoka2021}, and saliency maps \citep{Nusinovici2022}. In \citep{Poplin2018} it was shown that vascular regions in the retina are not only associated with age, but can also indicate cardiovascular risk, while perivascular surroundings reflect changes in Haemoglobin A1c levels. Interestingly, sex differences in the ocular fundus were found to be concentrated in the optic disc, vessels and macula. Retinal vessels have also been associated with retinal age estimation in \citep{Zhu2023}. For age estimation by macular OCT, whole layers of retina were found to be the most important for all age groups from 20 to 80 years \citep{Chueh2022, Shigueoka2021}. Retinal fundus photographs were also used to estimate mortality, with macula, optic disc, and retinal vessels manifesting the highest contribution \citep{Nusinovici2022}.

Tabular data are also used to estimate retinal age, albeit not very often, and usually represent various morphological metrics measured from retinal fundus photographs. Such metrics can be measured manually, or modern medical image segmentation approaches can be used. In this case, one relies on the other XAI approaches, for example, PFI \citep{Ma2021}. In \citep{Ma2021} the anterior chamber volume was identified as the most important metric for age estimation, negatively correlated with age, along with the absolute degree of anterior corneal astigmatism and corneal thickness parameters. 

\subsection{Circulatory and Cardiovascular Systems}\label{sec:XAI:subsec:cardiovascular}

\subsubsection{Age estimation with blood data}\label{sec:XAI:subsec:cardiovascular:blood}

Blood is an integral part of the cardiovascular system. It circulates within the human body, communicating with all systems, and it is an informative indicator of health status. The biochemical blood test is a common tool for assessing the human condition in clinical practice and is a rather sensitive indicator of many pathologies, including age-associated changes and diseases. Whole blood, its various products and indicators were among the first data types, which served as a basis for building models of human biological age estimation or the so-called ‘clocks’.

Blood biochemistry data stay in the focus of attention. These data always have a tabular format, representing a set of numerical values for each individual. In this regard, classical methods \citep{Sagers2020}, gradient ensemble approaches \citep{Wood2019}, and neural networks \citep{Mamoshina2018, Mamoshina2019a, Putin2016} are used to estimate age in this case, and XAI methods, such as PFI \citep{Mamoshina2018, Mamoshina2019a, Putin2016} and SHAP \citep{Wood2019}, are used to explain predictions. However, not all biochemical parameters and cell counts affect biological age equally. In \citep{Putin2016}, the levels of albumin, glucose, alkaline phosphatase, urea, and erythrocytes were found to be important markers of age in the constructed hematological clocks. When studying the ethnic specificities of age estimation, 5 markers, namely albumin, hemoglobin, urea, and glucose, were the most predictive for all the considered populations: Canadian, South Korean, and Eastern European \citep{Mamoshina2018}. A higher rate of aging in smokers was shown by hematological clocks in \citep{Mamoshina2019a}; it was also found that such cardiovascular risk indicators as high cholesterol ratio and fasting glucose significantly influenced age estimation in smokers. Glucose was also at the top by importance for both males and females \citep{Wood2019}, SHAP was also used to explain individual predictions, highlighting the contribution of levels of each individual feature to the resulting age estimation for each participant. In \citep{Sagers2020},  a built-in feature importance method for random forest showed that the ranking of features for age estimation was highly dependent on age range and highly correlated with sex and race/ethnicity.

Recently, there emerged a considerable interest in inflammatory profile data, also obtained from whole blood. It can characterize not only the status of the human immune system but also can be associated with the phenomenon of inflammaging - an increase in circulating inflammatory mediators with age. Immunological profile is characterized by tabular data (numerical values of cytokine levels) defining the choice of particular ML and XAI methods. The results on explaining inflammatory age models were published in \citep{Kalyakulina2023}. Here, by means of SHAP values, it was demonstrated that a crucial role in age estimation is played by levels of CXCL9, CD40LG, PDGFB in healthy participants and levels of CXCL9, IL6, CSF1 in patients with end-stage renal chronic disease. 

Epigenetic clocks that rely on DNA methylation data have gained extreme attention over the last decade. Whereas DNA methylation of any human tissue and/or organ can be considered, the whole blood methylation is the most commonly used for age estimation due to its low invasiveness. A number of technologies allow for obtaining DNA methylation data with different resolutions. Low resolution data are more frequently used in forensic applications, and high resolution epigenome-wide data in research. Like the other whole blood derived data, they are tabular.

The celebrated works proposing epigenetic clocks for both research \citep{Hannum2013, Horvath2013, Levine2018, Lu2019} and forensic applications \citep{Park2016, Zbiec-Piekarska2015} used linear models, in particular ElasticNet, which are easily explainable. Beside, classical models, which have a built-in functionality for determining the importance of all input features, have been used to construct epigenetic clocks \citep{Gao2020, Montesanto2020}. However, more recent approaches to epigenetic age estimation, to name variational autoencoders \citep{Levy2020}, DNNs \citep{deLimaCamillo2022}, tabular data-handling architectures like TabNet \citep{deLimaCamillo2022} required interpretations by XAI methods like SHAP \citep{Levy2020, deLimaCamillo2022} and DeepPINK \citep{deLimaCamillo2022}. Such approaches allowed for identifying meaningful relationships between the methylation of different genome regions, aging, and age-associated diseases. In particular, \citep{Levy2020} proposed MethylNet, one of the first models that beside accurate age estimation or disease classification, explained the obtained results using XAI. CpG sites significantly affecting age estimations differed significantly for participants younger than 44 years and older. This suggests a crossover in DNA methylation patterns change about a particular age, defining young and old age phenotypes. In \citep{deLimaCamillo2022} a pan-tissue epigenetic clock called AltumAge was proposed, supplied with a mechanism to explain predictions based on SHAP and DeepPINK. Although methylation data from various organs and tissues were used, special attention was paid specifically to blood DNA methylation and comparison with existing models. The individual most age-related features with different types of dependence (linear and nonlinear) were identified. It also suggested features associated with various diseases affecting age-related acceleration in blood methylation (HIV, Down Syndrome, autism, atherosclerosis).

Toxicological blood tests giving levels of various metabolites were also investigated as a basis for age estimation models. In \citep{Lassen2023}, metabolic profiles of drivers suspected of being under drug influence were examined and used for age estimation by ML techniques. SHAP analysis demonstrated that high levels of such age-related, age-associated diseases and stress biomarkers as acylcarnitines, cortisol, and benzoic acid contribute to age acceleration. At the same time, high levels of tryptophan pathway metabolites, serotonin, and kynurenate contributed to age deceleration. 

Another interesting type of data used for age estimation is circular RNA (circRNA) levels, which can also be obtained from human blood. It has a promising application in forensic science, especially in the case of fragmentary available data after the discovery of crime evidence, when it is necessary to determine the age of a victim or a criminal. Like all the previously discussed blood parameters, this one is numerical and the input data itself is tabular. Still, age estimation and model explanations based on circRNA data are not very common at the moment, with an exception of \citep{Wang2022}. VIM approach was taken there to determine the contribution of different circRNAs to age estimation in terms of the effect on model accuracy variation and standard deviations.

\subsubsection{Heart and arterial age estimation with image data}\label{sec:XAI:subsec:cardiovascular:image}

Cardiovascular age can be assessed by MRI or ultrasound images. In this case, one typically makes use of specialized neural network architectures (CNNs) for age estimation, and, accordingly, specialized XAI methods like attention maps. In \citep{Goallec2021} frames from cardiac MRI videos and cardiac MRI images were employed to estimate heart age, and attention maps suggested that mitral and tricuspid valves, aorta, and interventricular septum were the most important regions. In \citep{Goallec2021a} carotid ultrasound images were used to estimate arterial age, and attention maps showed that the carotid artery itself, as well as surrounding tissue and jugular veins made the highest contribution. Vascular images obtained by magnetic resonance angiography are also used in the assessment of cardiovascular age. It was shown in \citep{Nam2020} that the vascular regions along the cerebral arteries are most contributing to the cerebral vascular aging estimation result.

\subsubsection{Cardiovascular age estimation with signal data}\label{sec:XAI:subsec:cardiovascular:signal}

Signals (sequential data) can also be used in age estimation. ECG is a noninvasive test, extremely common in diagnostic applications. So ECG signals are the straightforward and most popular choice for cardiovascular age. One elegant and fairly simple option is to reduce the data to tabular. To do this, numerical metrics (usually durations and amplitudes) of different waves and complexes of ECG signal are calculated and used as an input. Based on it, classical and linear models with a built-in functional for determining feature importance for heart age estimation were used \citep{Attia2021, Lindow2022, Starc2012}. In \citep{Starc2012}, the normalized RR-interval variability (informative of arrhythmia) turned out to be the most important feature. A coherent result was obtained in \citep{Attia2021}, where the mean RR-interval duration, as well as the maximum amplitudes of the peaks R and S of the QRS complex were among the most important features for age estimation. In this paper, the authors considered two sets of features: human-derived features and neural network-derived features. Although most of the neural network-derived features were correlated with human-derived features, the models based on neural network-derived features performed better in terms of coefficient of determination. The authors suggested that neural network-derived features may reflect signal components that are extremely difficult for humans to recognize for various reasons (need for very high medical skills, inability to describe them in any natural language) or may represent more complex features compared to traditional methodology. In \citep{Lindow2022}, the heart age was calculated separately for men and women, and it was shown that the sets of important features in the two sexes are very similar and include, in particular, length of P wave, QT interval, and heart rate.

More modern and advanced methods allow the signals themselves to be used as input data for prediction models. In particular, image-based approaches can be adapted to unidimensional data (signals), in which case the approaches previously considered for images, such as saliency maps \citep{Lima2021}, can also be used to explain predictions. Additionally, it was demonstrated that different physicians can perform much worse, making significantly different estimations of individual age after visual inspection of ECG recordings. At the same time, saliency maps suggested that the highest contribution to the prediction is made by the low-frequency components of the ECG, which include, in particular, P and T waves. Models capable of handling biomedical signals, like ECG12Net, were also developed, and XAI methods, such as CAM \citep{Chang2022}, were applied to them. In result, the association of relative irregular baseline, as well as aVL leads in general with accelerated cardiac age for patients with coronary artery disease has been shown \citep{Chang2022}. 

Photoplethysmogram (PPG) that characterizes the pulse wave velocity and the filling of small vessels with blood is another type of signals widely used for a rapid and non-invasive assessment of cardiovascular health. As for ECG, one possibility is to convert a signal to tabular data. For this purpose, various temporal, amplitude, frequency metrics are calculated, which are used as input for ML models, in particular GBDTs \citep{Shin2022}. Although GBDT models have a built-in functionality for determining feature importance, it can benefit from a deeper analysis of global and local explainability by XAI approaches, like SHAP. In \citep{Shin2022}, the nasal PPG was analyzed to reveal the highest contribution to age estimation from the difference between incident wave peak and reflected wave peak amplitudes (that tends to decrease with age). As for ECG, modern methods like CNN, which take waveforms as input and use the Grad-Cam method of XAI to explain the predictions, were applied to PPG \citep{Shin2022a}. In essence, this approach evaluates the shape of individual signal elements and features it as an integral part of the data, taking part in the age estimation. In particular, \citep{Shin2022a} demonstrated that the waveform near the systolic peak contributes most to the characterization of vascular aging and this result is consistent across age groups.

\subsection{Respiratory System}\label{sec:XAI:subsec:respiratory}

\subsubsection{Age estimation with chest X-ray data}\label{sec:XAI:subsec:respiratory:xray}

Chest X-ray is a common medical procedure that is often performed to assess lung function and detect signs of various respiratory diseases. It can be used to assess not only the respiratory age of an individual but also the age of the structures adjacent to the lungs. As it was already pointed out, modern neural network architectures allow the direct use of X-ray images as an input for respiratory age estimation. XAI methods, such as saliency maps \citep{Karargyris2019} and Grad-CAM \citep{Ieki2022, Raghu2021}, are used for identifying the parts of the X-ray images most contributing to age estimation. In \citep{Karargyris2019}, different projections of chest X-rays were investigated, and such regions as the neck, clavicles, mediastinum, ascending aortic arch and spine were highlighted as the most important for age estimation. Interestingly, the important regions appeared to be age-dependent: lungs and bone and joint regions (clavicles and spine) were more significant only in younger participants. In \citep{Raghu2021}, it was found that the mediastinum, heart silhouette, and aortic protrusion dilate and become tortuous with age. Age estimation was also affected by such parts of the images as the diaphragm silhouette, upper mediastinum and lower neck, associated with age-related degenerative changes in the lower cervical spine. Aortic tortuosity and calcification, also found in \citep{Ieki2022}, have been shown to be associated with aging and development of atherosclerotic diseases. Such changes in the lungs, such as fibrosis, detected on X-rays, significantly increased the estimated age.

\subsection{Endocrine and Digestive Systems}\label{sec:XAI:subsec:endocrine}

\subsubsection{Abdominal age estimation with MRI data}\label{sec:XAI:subsec:endocrine:MRI}

Age-related changes affect all systems and organs of the human body, including abdominal organs. Since these organs are deep inside the body, MRI has become a practical tool for assessing their condition. As it was already pointed out, MRI images have a potential in estimating biological age for different organs, and attention maps or other suitable XAI methods can be useful in identifying important image regions. Currently, the studies on age estimation from abdominal MRI scans are quite limited. In particular, liver and pancreas MRI images were used for constructing for respective age models \citep{LeGoallec2022}. Attention maps built separately on liver and pancreatic MRI images highlighted common important areas of the abdomen, including the liver, stomach, spleen, as well as muscle, bone, and fatty tissue. This may reflect age-associated changes in the liver associated with inflammation, decreased blood flow, and decreased liver volume. Age-related changes in the pancreas included fatty degeneration and lobularity. Interestingly, the areas of MRI indicative of decelerated and normal aging were concentrated in the liver, whereas those related to accelerated aging were found in the pancreas and stomach.

\subsubsection{Age estimation with gut microbiome data}\label{sec:XAI:subsec:endocrine:gut}

The gut microbiome is a huge community, a complex and constantly evolving system. It is responsible for many body functions, including digestive, immune, and metabolic ones, and can also be informative of an individual's health status. The models for assessing age by the gut microbiome have been proposed, and in this case XAI methods have proved to be of particular importance. That is, they allow to estimate which microorganisms contribute most to age estimation and which composition of microbiota is associated with accelerated or decelerated aging. Gut microbiota data are quantitative estimates of microorganisms numbers, therefore, they are tabular. Accordingly, both built-in explainability methods for classical models \citep{Gopu2020, Huang2020, Shen2022} and specialized XAI methods, like ALE \citep{Galkin2020} and PFI \citep{Chen2022}, can suitably be applied. 

To cite some interesting results, it was shown that the gut microbiome patterns that are important for age estimation can be region specific \citep{Huang2020}. In particular, Bifidobacterium level contributed to aging clocks in the Chinese cohort, while Lachnospiraceae, Ruminococcaceae, and Clostridiaceae levels were generically important. In \citep{Gopu2020}, it was found that Haemophilus, Turicibacter, and Romboutsia groups were the most ones negatively correlated with age (i.e., an increase in their number is associated with a decrease in the predicted age), while Streptococcus and Propionibacterium were the most positively correlated ones (an increase in their number is associated with an increase in the predicted age). In \citep{Galkin2020}, the list of important features contained both those that have a positive effect on intestinal function (Bifidobacterium spp., Akkermansia muciniphila, Bacteroides spp.) and those having a negative effect (Escherichia coli, Campylobacter jejuni). Finegoldia magna, Bifidobacterium dentium, and Clostridium clostridioforme, as were shown in \citep{Chen2022}, had an abundance with age. Interestingly, Cellulosilyticum was abundant in the long-lived group \citep{Shen2022}.

\subsection{Skeletal and Muscular Systems}\label{sec:XAI:subsec:skeletal}

\subsubsection{Bone age assessment}\label{sec:XAI:subsec:skeletal:bone}

Estimation of bone age by hand X-rays is becoming widespread in clinical practice. They are most often used for two opposite age categories - in pediatrics and gerontology. During growth and development, particular parts of the child's hand bones develop at specific times, so X-rays assessment and comparing to age norm allows to detect and monitor the progression of genetic, endocrinological and other diseases. On the other hand, disorders in the bone structure are associated with many age-related diseases, and bone fragility frequently develops in elderly. Moreover, the extension of methods for bone age assessment to all age ranges can also be practical. Since bone age assessment relies on images, XAI methods, such as attention maps \citep{Lee2017, Wu2019}, CAM \citep{Bui2019, Zhao2018}, and detection of regions of interest (ROIs) corresponding to the most active neurons \citep{Spampinato2017}, are used to identify the regions of X-rays most important for the outcome.

In \citep{Spampinato2017} the authors proposed a CNN-based BoNet model developed specifically for processing X-rays images of hands. The most active neurons of this network highlighted radius and ulna, as well as tiny parts of carpal zones (at variance to the result based on the classical Tanner-Whitehouse method that highlights entire carpal zones). Interestingly, the regions highlighted as the most important for bone age estimation vary in different age groups \citep{Lee2017}. Attention maps for CNN showed that in prepuberty the model focuses on carpal bones and mid-distal phalanges, phalanges are most informative in early-mid and late-puberty, while in post puberty the wrist (where radius and ulna are close to each other) takes the lead. At the same time, no significant sex differences were found. However, a more detailed analysis in \citep{Zhao2018} using CAM showed that the metacarpal bones are important for predicting bone age in males, while for females the method focuses on a large number of hand bones, including caudal phalanges, metacarpal bones and carpal bones. Carpals have been shown to be important for infants and toddlers, while metacarpals and phalanges for older ages \citep{Wu2019}. Instructively, it was shown that noise can introduce uncertainty into the outcome of attention maps and choose irrelevant (mostly background) regions of the image.

An interesting approach was proposed in \citep{Bui2019}. Instead of using the X-ray image of the whole hand for bone age estimation, they limited it to six parts, selected by Tanner-Whitehouse (TW3) methods: dp3 (distal phalanx of the third finger), mp3 (middle phalanx of the third finger), pp3 (proximal phalanx of the third finger), mc1 (first metacarpal), ulna, radius. Attention maps were used for a more detailed analysis of the regions important for age assessment. Differences between age groups were also quite pronounced in this case, indicating the size, shape and degree of skeletal maturity change.

An important application of bone age estimation models is forensic science. Estimation of age-at-death (exact value or range) is an important step in the study of human remains, and pelvic bones are most commonly used for this purpose. Bone characteristics such as surface estimates, texture, and structure of different parts of the pelvis are usually considered as an input for ML models. Correspondingly, the data have a tabular structure, and both classical methods and DNNs are used to estimate bone age. In \citep{Koterova2018}, the built-in XAI techniques for classical methods highlighted such regions of the pubic symphysis as posterior plate, ventral plate, dorsal lip. They allowed for distinguishing between the samples under 30 years old, 30-40 years old, and over 40 years old. Some less frequent rule-based XAI methods were also employed in explaining age-at-death estimation in forensic applications \citep{Gamez-Granados2022}. Here, the articular face, dorsal plateau, and ventral margin demonstrated importance in age assessment for all age ranges. Upper symphysial extremity, bony nodule and lower symphysial extremity played a major role in young samples, while irregular porosity and ventral bevel took it over in older samples.

\subsubsection{Dental age estimation}\label{sec:XAI:subsec:skeletal:dental}

Teeth are rather resistant to negative environmental factors, but at the same time, they explicitly reflect age changes. Age estimation using dental images is used in archaeology, anthropology, forensic science and other applications where it is necessary to verify age or provide evidence that a person is a child or an adult. Dental information can be assessed in different forms, among which orthopantomography, panoramic radiological imaging of teeth and surrounding bone structures, is the most common. Modern DL approaches can handle such images as input, and the already described XAI methods are used to identify the most important parts of the image taken by the model for decision: DeepLIFT \citep{deBack2019}, Grad-CAM \citep{Atas2022, Guo2021, Kim2021, Sathyavathi2023, Vila-Blanco2020, Wallraff2021}. 

In \citep{deBack2019} the orthopantomograms of participants 5-25 years old were considered. It is interesting that for the youngest participants, the maxillary sinus added to the molars as the most informative region. At older ages, the nasal septum was found to be highly informative, together with the molars. The mandibular molars made the highest contribution to age estimation in \citep{Vila-Blanco2020} for samples younger than 25 years, in \citep{Wallraff2021} for samples 11-20 years old, in \citep{Sathyavathi2023} for samples 10-30 years old, and in \citep{Hou2021} for samples 0-93 years old. Interestingly, in \citep{Guo2021}, the deep CNN for age estimation focused on low-density regions on X-ray imaging, like dental pulp cavity, periodontal membrane, area between adjacent teeth and area between deciduous and permanent teeth. In \citep{Atas2022}, the gingival tissue and bone of the maxilla were found to be important for age estimation, in addition to the teeth.

In \cite{Kim2021} authors proposed to focus on the images of the first molars only for each individual (two first molars for each jaw - 4 images in total). It should be noted that instead of age regression, they addressed the classification of age ranges. It was shown that even for a particular tooth, the details important for age estimation differ between age groups. The first molar pulp was the most important for all the considered decadal age groups under 50 years (0-9 years, 10-19 years, 20-29 years, 30-39 years, 40-49 years) and over 60 years. Age range of 0-9 years was also characterized by the eruption degree of the second molar; the age range of 10-19 years was better characterized by the condition of the alveolar bone and maxillary sinus. At the age range of 20-29 years, the periapical area of the first molar came to the focus, at the age of 30-39 years - the interdental space between the first molar and the second molar, and at the age of 40-49 and 50-59 years - the interdental space and level of alveolar bone between the first molar and the second molar. For patients over the age of 60 years, the occlusal levels of the teeth proved to be important.

An interesting approach was proposed in \citep{Vila-Blanco2022}. The authors presented a special architecture of the CNN, which first performed segmentation of the panoramic image and highlighted the regions corresponding to each individual tooth. Then, based on these regions, a separate network was constructed which generated estimated per-tooth age distributions for each subject and a final prediction based on certain aggregation policies. For young subjects, it has been shown that the distributions for canines and premolars are best centered on real age.

At the same time, like for the images of other body structures, the panoramic images of teeth can also be recast in the tabular form. In this case, various numeric, ordinal, and categorical features describing each individual image are taken as input for ML models. To determine the most important features, both built-in feature importance methods \citep{Stepanovsky2017} and special methods, such as SHAP \citep{Lee2022, Patil2023}, are employed.

The work \citep{Stepanovsky2017} used unusual input: for each tooth, the numerical value of each feature was the average age over some representative population, reflecting the same type and degree of dental development. Only samples from 3 to 20 years old were considered. It was found that for males, the most important teeth for age estimation are all mandibular (lower jaw) molars, as well as the 2nd premolars, 2nd and 3rd molars of the maxilla (upper jaw). For females, the 1st premolars, 2nd molars on the mandible and central incisors, canines, 1st premolars, 1st and 2nd molars on the maxilla head the importance list. The work \citep{Lee2022} relied on more traditional metrics derived for panoramic radiographs, the sizes of certain teeth, interdental intervals, root and crown lengths. The most specific features for the young and elderly participants proved different. For the former, they were the distance between the mandibular canal and alveolar crest, tooth and pulp areas of the upper first molar, and pulp area of the lower first molar. For the latter, they were the number of teeth, the number of implants, the number of crown treatments, and the presence of periodontitis. In other words, the features for the younger group were mostly the characteristics of the first molars, and the features specific to the elderly refer more to the general condition of the teeth. In \citep{Patil2023}, where the authors focused on the images of the second and third molars, and even more specifically, on the values of the mesial and distal roots for each tooth. The length of the right side third molar mesial root proved to be the most important for the classification of all considered age ranges (12-25 years with division into 2, 3 and 5 equal groups).

\subsubsection{Muscular age estimation with gene expression data}\label{sec:XAI:subsec:skeletal:gene}

It is also possible to estimate an individual's age using more complex data, in particular gene expression profiles. They are numerical values of the expression levels of multiple genes, thus representing typical tabular data. Obtaining such data is more costly and time-consuming, which limits the use in age estimation. Gene profiles can be obtained for different body systems, in particular, for skeletal muscles, as in \citep{Mamoshina2018a}. In this work, the authors considered classical methods with built-in functionality for determining the importance of individual features; they also used the Borda count algorithm to combine the rankings of the features for different models. Among the most important genes, authors found those known to be therapeutic targets for many drugs, as well as genes related to skeletal muscle relaxation. The authors suggested that this result may be important in the development of neuromuscular damage therapy.

\subsection{Integumentary System}\label{sec:XAI:subsec:integumenary}

\subsubsection{Age estimation with skin microbiome data}\label{sec:XAI:subsec:integumenary:skin}

Skin has its own microbiome that mounts a complex system under contact with the environment. It changes with age and can therefore also be used to characterize the health status. The models estimate age by skin microbiome composition, and the respective XAI methods match the abundance of different groups of microorganisms with accelerated or decelerated aging. As for the gut microbiome the data are tabular, and, accordingly, both built-in explainability methods for classical models \citep{Huang2020} and specialized XAI methods, like SHAP \citep{Carrieri2021}, are applicable. 

In \citep{Huang2020} it was found that the age-associated composition of the skin microbiome differs in males and females; the differences between the forehead and palm microbiomes were pointed out. Mycoplasma, Enterobacteriaceae, and Pasteurellaceae groups, involved in age-dependent changes in physiological skin characteristics, such as sebum production and dryness, were negatively correlated with age. In \citep{Carrieri2021}, it was suggested that different microbiome families manifest high importance in the age model for young and old age groups. In particular, it reported a decrease in the relative abundance of Propionibacterium with age, associated with the decrease in sebum secretion. Alloprevotella, Granulicatella, Gemella and Lactobacillus families were also found among the features specific for the younger age. Bacillus, Bacteroides, Pseudomonas, and Bergeyella families were identified as the most important features for age estimation of elderly people.

\subsubsection{Epidermal age estimation with gene expression data}\label{sec:XAI:subsec:integumenary:gene}

Gene expression profiles, as discussed earlier, are tabular data representing numerical values of gene expression levels. Transcriptome profiles for epidermis were investigated in \citep{Holzscheck2021}. The authors calculated pathway ranking based on the correlation of activations of intermediate neural network neurons with chronological age, which is an uncommon approach. The pathways responsible for p53 and TNFa/NFkB signaling, as well as responses to ultraviolet radiation and interferon gamma, were found to be the most significantly associated with age.

\subsection{Complex Data}\label{sec:XAI:subsec:aggregated}

Not only biomarkers localized to a single body system can be used for age prediction. Aging is a complex process affecting the whole organism, so age-related changes can develop at different rates and intensities for different organism structures. In this case, more complex biomarkers affecting multiple systems of the human body can be considered. This potentially allows the identification of higher-level correlations between aging patterns between these systems.

\subsubsection{Age estimation with whole-body MRI data}\label{sec:XAI:subsec:aggregated:MRI}

Whole-body MRIs survey the human body from the neck to the knees, and whole-body X-rays make it from head to feet, depicting internal organs as well as muscle and fat tissue distributions. Being less detailed than images of individual organs (e.g., the brain), they provide a more comprehensive assessment, and have a potential to yield integrative clocks that characterize the aging status of many body systems, possibly each having individual pace, and to infer interactions between them. MRI/X-ray images are most suitably addressed by CNNs for age estimation, whereas saliency maps \citep{Langner2020} and Grad-RAM \citep{Goallec2021b} are used to explain their predictions.

In \citep{Langner2020}, MRI image sets for each individual were combined into two types of images, based on the water and fat signals in two projections. The most active regions on the saliency maps included the knee joint, the aortic arch, the area covering the heart, the surrounding tissues, and part of the lungs. In younger participants, the knees, together with the contour of the tibia and the outer edge of the calf muscle, almost always appeared to be important, unlike the thoracic region. In \citep{Goallec2021b}, single projection X-ray images were used to construct the age predictor. The most important areas on the attention maps included neck, upper body, hips, and knees. Further on, the maps were also constructed specifically for selected body parts: they demonstrated importance of the lumbar region for X-ray images of the spine in the sagittal projection; the greater trochanter of the femur and the joint itself for the hip joint; and the thigh bone, the tibia, and the joint itself for the knee.

\subsubsection{Face age estimation}\label{sec:XAI:subsec:aggregated:face}

Estimating age from facial images is frequently addressed in automatic facial analysis. Beside a chronological age, a visual age is influenced by the health status, environmental exposures, and many other factors. Aging signatures can be exposed on the face unevenly, get manifested at different levels and progress at different pace. In result, while obtaining data is simply taking a snapshot, facial age estimation is a difficult task. The models working specifically with images as input (mainly CNNs and their modifications) were first applied here. 

In \citep{Agustsson2017} a neural network architecture capable of analyzing facial images was proposed. It included pre-processing, aligning and predicting two types of age - real and apparent (perceived by human observers). Sensitivity maps for individual pixels in relation to the estimated age allowed the authors to determine which parts of the image the models rely on when making a decision. It appeared that sensitivity areas were changing with age: the forehead and the space between the eyes were important for young people, for middle-aged people the important areas were relatively uniformly distributed across the face, while for senior individuals the chin and the neck areas took the lead. In \citep{Gao2018}, activation (score) maps for different age groups (children 0-3 years old, adults 20-35 years old, seniors 65-100 years old) also highlighted different areas. The results for the first two groups agreed with \citep{Agustsson2017}: for infants, the eyes were the most important, for adults the eyes, nose, and mouth were decisive for age estimation. For seniors, the key areas included forehead, eyebrows, eyes, and nose. In \citep{Abdolrashidi2020}, edge patterns around facial parts as well as wrinkles proved central for predicting sex and age. \citep{Letzgus2022} confirmed the importance of the eyes for age prediction in infants, the central part of the face in adults (mainly eyes, nose, and mouth), and the whole face in seniors (with a particular contribution of wrinkles).

\subsubsection{Pan-tissue epigenetic and transcriptomic clocks}\label{sec:XAI:subsec:aggregated:clocks}

Age predictor models based on omics data, obtained from a specific tissue (most frequently, the whole blood) can be made more universal to accept data derived from different tissues. Such models are called pan-tissue clocks. The celebrated pan-tissue epigenetic clocks proposed by Horvath \citep{Horvath2013} were based on the ElasticNet linear model and thus are easily interpreted. The more recent AltumAge model presented in \citep{deLimaCamillo2022} employed a DNN architecture, and maked use of SHAP and DeepPINK approaches to explain predictions. In particular, it demonstrated age acceleration in brain samples from patients with autism and multiple sclerosis, in liver samples from patients with non-alcoholic fatty liver disease, and in pancreatic samples from patients with type 2 diabetes. XAI approaches also helped to identify the relationship between chromatin states and age predictions for a number of tissues. 

In \citep{Shokhirev2021}, the proposed age predictor based on transcriptomic data revealed the tissue-specificity and sex-specificity of feature rankings in ensemble models. The authors found age-associated genes shared by retina, brain, blood, heart and bones, constructing an age estimator based on transcriptomic data from the corresponding tissue.

The multimodal aging clock proposed in \citep{Urban2023} incorporated both methylation and transcriptomic data in a tissue-agnostic fashion. SHAP was used to identify the most important genes that correspond to pathways associated with aging and age-related diseases: tRNA processing in mitochondrion, amino acid transport across plasma membrane, suppression of apoptosis, vasopressin-like receptors, highly sodium permeable postsynaptic acetylcholine nicotinic receptors, cytosolic sulfonation of small molecules. The authors also revealed the association of the obtained genes with drug targets for idiopathic pulmonary fibrosis, chronic obstructive pulmonary disease, Parkinson's disease and heart failure.

\subsubsection{Age estimation with medical records data}\label{sec:XAI:subsec:aggregated:record}

Medical records can include clinical history, anthropometric measures, laboratory tests (blood, urine, feces), physical examination results, and many other characteristics. They have a strong potential for a comprehensive assessment of a person's health status and constructing informative clocks. These data are usually given in the tabular form with categorical, ordinal, and continuous values. Feature rating for classical models \citep{Yang2022}, PFI \citep{Bae2021} and SHAP \citep{Bernard2023} are used to determine the most important features among the input set and to explain model predictions.

In \citep{Bae2021} different types of models for age estimation based on medical record data were considered: linear and polynomial models, ensemble models and DNNs. PFI scores showed that levels of creatinine and aspartate aminotransferase, as well as waist circumference were among the most important features for all models. Sex, body mass index, lean body mass, lactate dehydrogenase and blood urea nitrogen levels consistently manifested high importance as well. Indexes characterizing the body constitution were also at the top. In \citep{Yang2022} the features ranking for an ensemble of classical models assigned the highest score to the values of diastolic and systolic blood pressure, height, sex, and platelet content. It was also shown that body shape index and waist-to-height ratio are associated with predicted age and serve health risk indicators. The physiological age model proposed in \citep{Bernard2023} along with SHAP analysis suggested that the parameters related to metabolism, nitrogen (uric metabolites and creatinine), carbon (glycohemoglobin, triglycerides, and glucose), and liver function (albumin, ALT, and GGT) contribute the most to age estimation. The authors also found a threshold for glycohemoglobin levels that distinguishes between younger and older samples.

In addition, we refer to mortality risk prediction models that are based on the same or similar input variables. XAI approaches, most commonly SHAP, are applicable here too \citep{Qiu2022a, Qiu2022, Thorsen-Meyer2020}. Along these lines \cite{Qiu2022a} demonstrated that red cell distribution width, serum albumin, arm circumference, platelet count, and serum chloride levels have the highest impact on 5-year mortality. Some of these parameters (red cell distribution width, serum albumin) had been known before as markers of mortality risk, while others (platelet count, serum chloride levels) were discovered for the first time. Further on, the importance of red cell distribution width was found to increase from 1-year to 10-year mortality, and the importance of serum albumin on the opposite. In \citep{Qiu2022} it was shown that cystatin C, smoking status, history for chronic and cancer diseases dominate the estimate of the all-cause mortality and neoplasm-cause mortality in women aged 65 years. An application of such mortality predictor models was described in \citep{Thorsen-Meyer2020}, which predicted 90-day mortality for patients in intensive care units. The most important parameters in this case were age at admission, heart rate, surgical intervention, blood pressure, blood oxygen saturation, Glasgow Coma Scale, temperature, length of stay in hospital before admission to the intensive care unit. Such models are particularly important for proper allocation of hospital workload and timely assessment of the risks of possible health deterioration.

\section{Conclusion}\label{sec:conclusion}

XAI is a powerful tool for interpreting the results of complex models, which is crucial in sensitive applications, such as healthcare. Such models must not only demonstrate a good performance, but also have to allow experts to validate the results of their decisions. This allows for monitoring the behavior of the model, interpreting correct answers and sources of errors, making corrections to the model “on the fly”, and as a result, ensuring trustworthy AI. We would like to stress that interpretability is an aspect of reliability that becomes a golden standard for all modern AI systems, which are usually black boxes with a non-transparent decision-making process. The review suggests that the development of XAI approaches and their application to age prediction models drive in the top gear.

\begin{figure}
	\centering
	\includegraphics[width=0.99\textwidth]{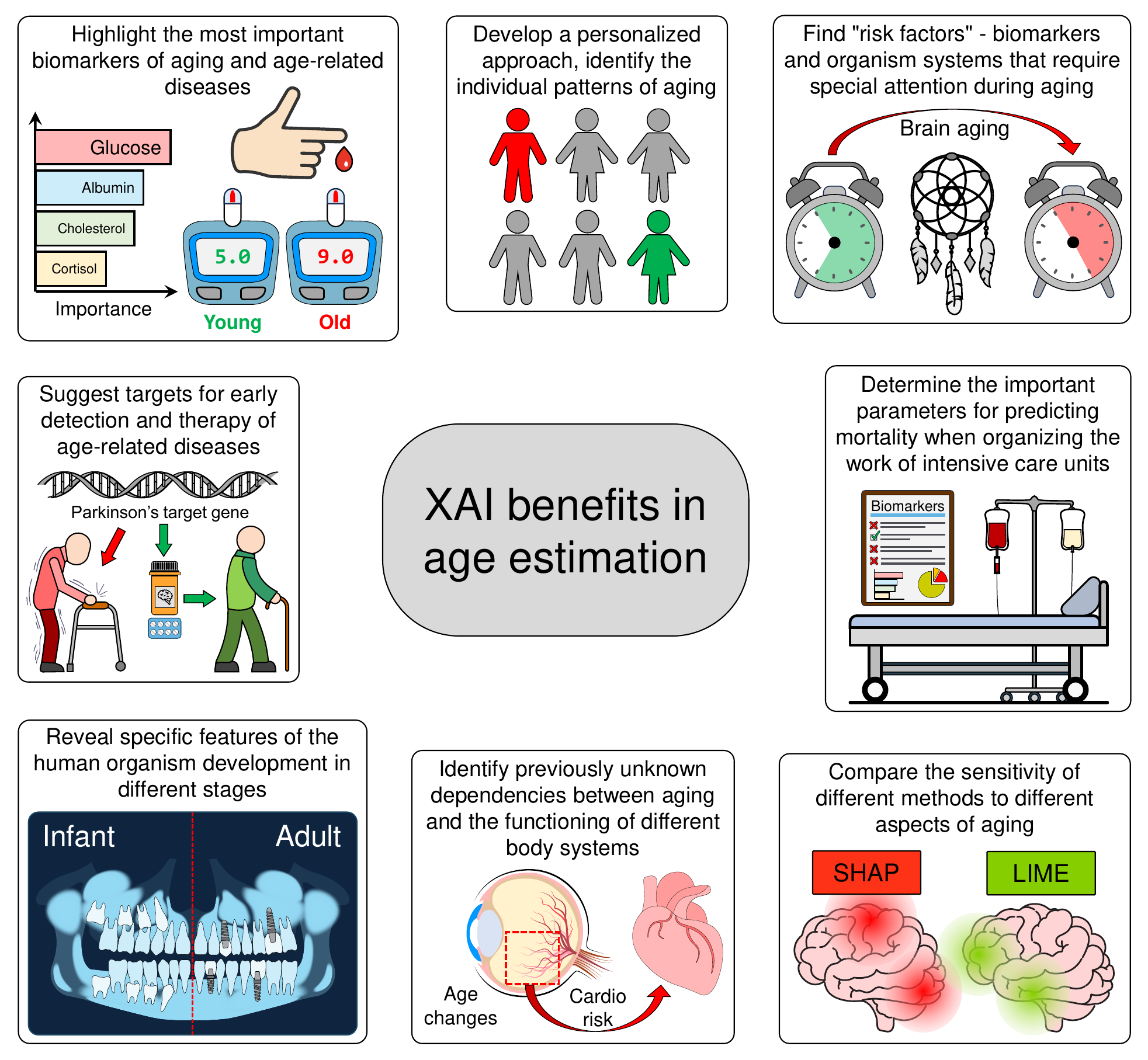}
	\caption{The main benefits that XAI provides in the age estimation task, with schematic examples for each point.}
	\label{fig:benefit}
\end{figure}

The application of XAI approaches to the biological age prediction task hits several targets, cf. Figure \ref{fig:benefit}. One of the most significant and obvious achievements is identifying the most important biomarkers of aging and age-associated diseases, while keeping the benefits of the high performance of “black-box” AI models. In result, explainable clocks and age-associated parameters for a broad spectrum of body systems are becoming available. XAI is also helpful in deriving ‘risk factors’ from AI-based age estimation and mortality prediction models. Such early predictors can serve as ‘red flags’ before serious symptoms emerge and the model explainability can pave the way to recommending anti-age corrections. Next, the majority of XAI methods provide explanations for each specific sample, identifying the individual signatures that made the highest impact on the final prediction for that particular person. This opens the door for personalized preventive approaches and therapies. XAI methods have a clear potential in identifying the early signatures of age-related diseases and the targets for the early correction and treatment. Another important aspect of XAI for public health is related to support the organization of intensive care units. Determining the most important features in predicting mortality will allow the proper allocation of healthcare facility resources and provide the most efficient care. XAI also gives much promise in the fundamental research, suggesting the previously unknown dependencies between different body systems in the aging process, and the comparison of the sensitivity of different approaches to different aspects of aging.

In the horizon of several years, XAI approaches are highly likely to become ubiquitous in the aging research, development of various clocks, biomarkers of aging and age-related diseases, with the strong emphasis on determining individual aging status, personalized recommendations, corrections and therapies.

\section*{Abbreviations}

AI - Artificial Intelligence; ALE - Accumulated Local Effects; ALT - ALanine Transaminase; CAM - Class Activation Mapping; CNN - Convolutional Neural Network; CSF - CerebroSpinal Fluid; DeepLIFT - Deep Learning Important FeaTures; DeepPINK - Deep feature selection using Paired-Input Nonlinear Knockoffs; DL - Deep Learning; DNA - DeoxyriboNucleic Acid; DNN - Deep Neural Network; ECG - Electrocardiogram; EEG - Electroencephalogram; GBDT - Gradient-Boosted Decision Tree; GGT - Gamma-Glutamyl Transferase; Grad-CAM - Gradient-weighted Class Activation Mapping; HIV - Human Immunodeficiency Virus; LIME - Local Interpretable Model-agnostic Explanations; ML - Machine Learning; MRI - Magnetic Resonance Imaging; OCT - Optical Coherence Tomography; PDP - Partial Dependence Plot; PFI - Permutation Feature Importance; PPG - PhotoPlethysmoGram; REM - Rapid Eye Movement; RNA - RiboNucleic Acid; ROI - Region Of Interest; SHAP - Shapley Additive exPlanations; VIM - Variable Importance Measure; XAI - eXplainable Artificial Intelligence.

\bibliographystyle{plainnat}
\bibliography{arxiv_preprint}  






\end{document}